\documentclass[11pt]{article}
\pdfoutput=1

\usepackage{acl2012}
\usepackage{times}
\usepackage{latexsym}
\usepackage{amsmath}
\usepackage{multirow}
\usepackage{stmaryrd}
\usepackage{url}
\usepackage{color}
\usepackage{graphicx}
\usepackage[utf8]{inputenc}
\usepackage[T1]{fontenc}
\usepackage{caption}
\usepackage{float}

\newcommand{\malopa}{\textsc{MaLOPa}}
\newcommand{\ignore}[1]{}

\definecolor{mygreen}{rgb}{0.1, 0.6, 0.1}
\definecolor{mygrey}{rgb}{0.6, 0.6, 0.6}

\newenvironment{itemizesquish}{\begin{list}{\labelitemi}{\setlength{\itemsep}{0em}\setlength{\labelwidth}{0.5em}\setlength{\leftmargin}{\labelwidth}\addtolength{\leftmargin}{\labelsep}}}{\end{list}}
\title{Many Languages, One Parser}

\author{Waleed Ammar$^{\diamondsuit}$ ~ George Mulcaire$^{\heartsuit}$ ~ Miguel Ballesteros$^{\spadesuit\diamondsuit}$ ~ Chris Dyer$^{\diamondsuit}$ ~ Noah A. Smith$^{\heartsuit}$\\
$^\diamondsuit$School of Computer Science, Carnegie Mellon University, Pittsburgh, PA, USA \\
$^{\heartsuit}$Computer Science \& Engineering, University of Washington, Seattle, WA, USA\\
$^\spadesuit$NLP Group, Pompeu Fabra University, Barcelona, Spain \\
{\tt wammar@cs.cmu.edu, gmulc@uw.edu, miguel.ballesteros@upf.edu} \\ {\tt cdyer@cs.cmu.edu, nasmith@cs.washington.edu}
}

\date{}

\begin{document}
\maketitle
\begin{abstract}
We train one multilingual model for dependency parsing and use it to parse sentences in several languages.
The parsing model uses (i)~multilingual word clusters and embeddings; (ii)~token-level language information; and (iii) language-specific features (fine-grained POS tags). This input representation enables the parser not only to parse effectively in multiple languages, but also to generalize across languages based on linguistic universals and typological similarities, making it more effective to learn from limited annotations.
Our parser's performance compares favorably to strong baselines in a range of data scenarios, including when the target language has a large treebank, a small treebank, or no treebank for training.
\end{abstract}

\section{Introduction}
\label{sec:introduction}

Developing tools for processing many languages has long been an important goal in NLP~\cite{rosner:88,heid:89},\footnote{As of 2007, the total number of native speakers of the hundred most popular languages only accounts for 85\% of the world's population \cite{native:07}.}
but it was only when statistical methods became standard that massively multilingual NLP became economical.
The mainstream approach for multilingual NLP is to design language-specific  models.
For each language of interest, the resources necessary for training the model are obtained (or created), and separate parameters are fit for each language separately.
This approach is simple and grants the flexibility of customizing the model and features to the needs of each language, but it is suboptimal for theoretical and practical reasons.
Theoretically, the study of linguistic typology tells us that many languages share morphological, phonological, and syntactic phenomena \cite{bender:11}; therefore, the mainstream approach misses an opportunity to exploit relevant supervision from typologically related languages.
Practically, it is inconvenient to deploy or distribute NLP tools that are customized for many different languages because, for each language of interest, we need to configure, train, tune, monitor, and occasionally update the model.
Furthermore, code-switching or code-mixing (mixing more than one language in the same discourse), which is pervasive in some genres, in particular social media, presents a challenge for monolingually-trained NLP models \cite{barman:14}.\footnote{While our parser can be used to parse input with code-switching, we have not evaluated this capability due to the lack of appropriate data.}

In parsing, the availability of homogeneous syntactic dependency annotations in many languages \cite{mcdonald:13,universal:v1_0,universal:v1_1,universal:v1_2} has created an opportunity to develop a parser that is capable of parsing sentences in multiple languages, addressing these theoretical and practical concerns.\footnote{Although multilingual dependency treebanks have been available for a decade via the 2006 and 2007 CoNLL shared tasks \cite{buchholz:06,nivre:07}, the treebank of each language was annotated independently and with its own annotation conventions.}
A multilingual parser can potentially replace an array of language-specific monolingually-trained parsers (for languages with a large treebank).
The same approach has been used in low-resource scenarios (with no treebank or a small treebank in the target language), where indirect supervision from auxiliary languages improves the parsing quality \cite{cohen:11,mcdonald:11,zhang:15,duong:15b,duong:15,guo:16}, but these models may sacrifice accuracy on source languages with a large treebank.
In this paper, we describe a model that works well for both low-resource and high-resource scenarios.

We propose a parsing architecture that takes as input sentences in several languages,\footnote{We discuss data requirements in the next section.} optionally predicting the part-of-speech (POS) tags and input language.
The parser is trained on the union of available universal dependency annotations in different languages.
Our approach integrates and critically relies on several recent developments related to dependency parsing: universal POS tagsets \cite{petrov:12}, cross-lingual word clusters \cite{tackstrom:12}, selective sharing \cite{naseem:12}, universal dependency annotations \cite{mcdonald:13,universal:v1_0,universal:v1_1,universal:v1_2}, advances in neural network architectures \cite{chen:14,dyer:15}, and multilingual word embeddings \cite{gardner:15,guo:16,ammar:16}.
We show that our parser compares favorably to strong baselines trained on the same treebanks in three data scenarios: when the target language has a large treebank (Table \ref{tab:with}), a small treebank (Table \ref{tab:small_treebank}), or no treebank (Table \ref{tab:without}).
Our parser is publicly available.\footnote{\url{https://github.com/clab/language-universal-parser}}

\begin{table*}[!]
\setlength{\tabcolsep}{.375em}
\centering
\scriptsize
\begin{tabular}{ll|rrrrrrr}
														&             & German (de) & English (en)  & Spanish (es) & French (fr) & Italian (it) & Portuguese (pt) & Swedish (sv) \\ \hline
\multicolumn{1}{l|}{\multirow{3}{*}{UDT 2}}				& train       & 14118 (264906)  & 39832 (950028)   & 14138 (375180)   & 14511 (351233)  & 6389 (149145)    & 9600 (239012)       & 4447 (66631)     \\ \cline{2-9} 
\multicolumn{1}{l|}{}										& dev.         & 801 (12215)     & 1703 (40117)     & 1579 (40950)     & 1620 (38328)    & 399 (9541)       & 1211 (29873)        & 493 (9312)       \\ \cline{2-9} 
\multicolumn{1}{l|}{}										& test        & 1001 (16339)    & 2416 (56684)     & 300 (8295)       & 300 (6950)      & 400 (9187)       & 1205 (29438)        & 1219 (20376)     \\ \hline
\multicolumn{1}{l|}{\multirow{4}{*}{UD 1.2}}        		& train       & 14118 (269626)  & 12543 (204586)   & 14187 (382436)   & 14552 (355811)  & 11699 (249307)   & 8800 (201845)       & 4303 (66645)     \\ \cline{2-9} 
\multicolumn{1}{l|}{}										& dev.         & 799 (12512)     & 2002 (25148)     & 1552 (41975)     & 1596 (39869)    & 489 (11656)      & 271 (4833)          & 504 (9797)       \\ \cline{2-9} 
\multicolumn{1}{l|}{}										& test        & 977 (16537)     & 2077 (25096)     & 274 (8128)       & 298 (7210)      & 489 (11719)      & 288 (5867)          & 1219 (20377)     \\ \cline{2-9} 
\multicolumn{1}{l|}{}										& tags & -               & 50               & -                & -               & 36               & 866                 & 134              \\ \hline
\end{tabular}
\caption{Number of sentences (tokens) in each treebank split in Universal Dependency Treebanks  (UDT) version 2.0  and Universal Dependencies version (UD) 1.2 for the languages we experiment with. The last row gives the number of unique language-specific fine-grained POS tags used in a treebank.}
\label{tab:treebanks}
\end{table*}

\section{Overview}

Our goal is to train a dependency parser for a set of \underline{t}arget \underline{l}anguages ${L}^t$, given universal dependency annotations in a set of \underline{s}ource \underline{l}anguages ${L}^s$.
Ideally, we would like to have training data in all target languages (i.e., $L^t \subseteq L^s$), but we are also interested in the case where the sets of source and target languages are disjoint (i.e., $L^t \cap L^s = \emptyset$).
When all languages in $L^t$ have a large treebank, the mainstream approach has been to train one monolingual parser per target language and route sentences of a given language to the corresponding parser at test time.
In contrast, our approach is to train one parsing model with the union of treebanks in $L^s$, then use this single trained model to parse text in any language in $L^t$, hence the name ``\underline{Ma}ny \underline{L}anguages, \underline{O}ne \underline{Pa}rser'' (\malopa).
\malopa~strikes a balance between:
(1) enabling cross-lingual model transfer via language-invariant input representations; i.e., coarse POS tags, multilingual word embeddings and multilingual word clusters, and
(2) tweaking the behavior of the parser depending on the current input language via language-specific representations; i.e., fine-grained POS tags and language embeddings.

In addition to universal dependency annotations in source languages (see Table \ref{tab:treebanks}), we use the following data resources for each language in ${L} = {L}^t \cup {L}^s$:
\begin{itemizesquish}
\item universal POS annotations for training a POS tagger,\footnote{See \S\ref{sec:joint} for details.}
\item a bilingual dictionary with another language in $L$ for adding cross-lingual lexical information,\footnote{Our best results make use of this resource. We require that all languages in $L$ are (transitively) connected. The bilingual dictionaries we used are based on unsupervised word alignments of parallel corpora, as described in \newcite{guo:16}. See \S\ref{sec:embeddings} for details.}
\item language typology information,\footnote{See \S\ref{sec:lang} for details.}
\item language-specific POS annotations,\footnote{Our best results make use of this resource. See \S\ref{sec:fine_grained} for details.} and
\item a monolingual corpus.\footnote{This is only used for training word embeddings with `multiCCA,' `multiCluster' and `translation-invariance' methods in Table \ref{tab:multilingual_embeddings}. We do not use this resource when we compare to previous work.}
\end{itemizesquish}

Novel contributions of this paper include: (i) using one parser instead of an array of monolingually-trained parsers without sacrificing accuracy on languages with a large treebank, (ii) an effective neural network architecture for using language embeddings to improve multilingual parsing, and (iii) a study of how automatic language identification affects the performance of a multilingual dependency parser.

While not the primary focus of this paper, we also show that a variant of our parser outperforms previous work on multi-source cross-lingual parsing in low resource scenarios, where languages in $L^t$ have a small treebank (see Table \ref{tab:small_treebank}) or where $L^t \cap L^s = \emptyset$ (see Table \ref{tab:without}).
In the small treebank setup with 3,000 token annotations, we show that our parser consistently outperforms a strong monolingual baseline with 5.7 absolute LAS (labeled attachment score) points per language, on average.

\section{Parsing Model}
\label{sec:model}

Recent advances suggest that recurrent neural networks, especially long short-term memory (LSTM) architectures, are capable of learning useful representations for modeling problems of sequential nature  \cite{graves:13,sutskever:14}.
In this section, we describe our language-universal parser, which extends the stack LSTM (S-LSTM) parser of \newcite{dyer:15}.

\subsection{Transition-based Parsing with S-LSTMs}
\label{sec:slstm}

This section briefly reviews Dyer et al.'s S-LSTM parser, which we modify in the following sections.
The core parser can be understood as the sequential manipulation of three data structures:
\begin{itemizesquish}
\item a buffer (from which we read the token sequence),  
\item a stack (which contains partially-built parse trees), and 
\item a list of actions previously taken by the parser. 
\end{itemizesquish}
The parser uses the arc-standard transition system \cite{nivre:04}.\footnote{In a preprocessing step, we transform nonprojective trees in the training treebanks to pseudo-projective trees using the ``baseline'' scheme in \cite{nivre:05}. We evaluate against the original nonprojective test set.} 
At each timestep $t$, a transition action is applied that alters these data structures according to Table~\ref{fig:parser}.

\begin{table*}[!]
\centering
\begin{tabular}{r|r||l|c||r|r}
\textbf{Stack}$_t$ & \textbf{Buffer}$_t$ & \textbf{Action} & \textbf{Dependency} & \textbf{Stack}$_{t+1}$ & \textbf{Buffer}$_{t+1}$ \\
\hline
$u,v,S$ & $B$  &$\textsc{reduce-right}(r)$ & $u \stackrel{\scriptsize{r}}{\rightarrow} v$ & $u,S$ & $B$ \\
$u,v,S$ & $B$ & $\textsc{reduce-left}(r)$ & $u \stackrel{\scriptsize{r}}{\leftarrow} v$ & $v,S$ & $B$ \\
$S$ & $u,B$ & \textsc{shift} & ---  & $u,S$ & $B$
\end{tabular}
\caption{\label{fig:parser} Parser transitions indicating the action applied to the stack and buffer at time $t$ and the resulting stack and buffer at time $t+1$.}
\end{table*}

Along with the discrete transitions of the arc-standard system, the parser computes vector representations for the buffer, stack and list of actions at time step $t$ denoted $\mathbf{b}_t$, $\mathbf{s}_t$, and $\mathbf{a}_t$, respectively.\footnote{A stack-LSTM module is used to compute the vector representation for each data structure, as detailed in \newcite{dyer:15}.}
The parser state at time $t$ is given by:
\begin{equation}
\mathbf{p}_t = \max\left\{ \boldsymbol{0}, \mathbf{W}[\mathbf{s}_t; \mathbf{b}_t; \mathbf{a}_t] + \mathbf{W}_{\text{bias}}\right\}
\label{eq:parser_state}
\end{equation}
where the matrix $\mathbf{W}$ and the vector $\mathbf{W}_{\text{bias}}$ are learned parameters. The matrix $\mathbf{W}$ is multiplied by the vector $[\mathbf{s}_t; \mathbf{b}_t; \mathbf{a}_t]$ created by the concatenation of $\mathbf{s}_t, \mathbf{b}_t, \mathbf{a}_t$.
The parser state $\mathbf{p}_t$ is then used to define a categorical distribution over possible next actions $z$:\footnote{The total number of actions is $1 + 2 \times$ the number of unique dependency labels in the treebank used for training, but we only consider actions which meet the arc-standard preconditions in Fig.~\ref{fig:parser}.}
\begin{equation}
p(z \mid \mathbf{p}_t) = \frac{\exp \left( \mathbf{g}_{z}^{\top} \mathbf{p}_t + q_{z} \right)}{\sum_{z'} \exp \left( \mathbf{g}_{z'}^{\top} \mathbf{p}_t + q_{z'} \right)}
\label{eq:action_prob}
\end{equation}
where $\mathbf{g}_z$ and $q_z$ are parameters associated with action $z$.
The selected action is then used to update the buffer, stack and list of actions, and to compute $\mathbf{b}_{t+1}$, $\mathbf{s}_{t+1}$ and $\mathbf{a}_{t+1}$ accordingly.

The model is trained to maximize the log-likelihood of correct actions.
At test time, the parser greedily chooses the most probable action in every time step until a complete parse tree is produced.

The following sections describe our extensions of the core parser.
More details about the core parser can be found in \newcite{dyer:15}.

\subsection{Token Representations}
The vector representations of input tokens feed into the stack-LSTM modules of the buffer and the stack.
For monolingual parsing, we represent each token by concatenating the following vectors:
\begin{itemizesquish}
\item a fixed, pretrained embedding of the word type,
\item a learned embedding of the word type, 
\item a learned embedding of the Brown cluster,
\item a learned embedding of the fine-grained POS tag,
\item a learned embedding of the coarse POS tag.
\end{itemizesquish}

For multilingual parsing with \malopa, we start with a simple delexicalized model where the token representation only consists of learned embeddings of coarse POS tags, which are shared across all languages to enable model transfer.
In the following subsections, we enhance the token representation in \malopa~to include lexical embeddings, language embeddings, and fine-grained POS embeddings.

\subsection{Lexical Embeddings}
\label{sec:embeddings}

Previous work has shown that sacrificing lexical features amounts to a substantial decrease in the performance of a dependency parser \cite{cohen:11,tackstrom:12,tiedemann:15,guo:15}.  
Therefore, we extend the token representation in \malopa~by concatenating learned embeddings of multilingual word clusters, and pretrained multilingual embeddings of word types.

\paragraph{Multilingual Brown clusters.} Before training the parser, we estimate Brown clusters of English words and project them via word alignments to words in other languages. This is similar to the `projected clusters' method in \newcite{tackstrom:12}.
To go from Brown clusters to embeddings, we ignore the hierarchy within Brown clusters and assign a unique parameter vector to each cluster.

\paragraph{Multilingual word embeddings.} We also use Guo et al.'s (2016) \nocite{guo:16} `robust projection' method to pretrain multilingual word embeddings.
The first step in `robust projection' is to learn embeddings for English words using the skip-gram model \cite{mikolov:13}.
Then, we compute an embedding of non-English words as the weighted average of English word embeddings, using word alignment probabilities as weights.
The last step computes an embedding of non-English words which are not aligned to any English words by averaging the embeddings of all words within an edit distance of 1 in the same language.
We experiment with two other methods---`multiCCA' and `multiCluster,' both proposed by \newcite{ammar:16}---for pretraining multilingual word embeddings in \S\ref{sec:with}.
`MultiCCA' uses a linear operator to project pretrained monolingual embeddings in each language (except English) to the vector space of pretrained English word embeddings, while `multiCluster' uses the same embedding for translationally-equivalent words in different languages.
The results in Table \ref{tab:multilingual_embeddings} illustrate that the three methods perform similarly on this task.

\subsection{Language Embeddings}
\label{sec:lang}

While many languages, especially ones that belong to the same family, exhibit \textit{some} similar syntactic phenomena (e.g., all languages have subjects, verbs, and objects), substantial syntactic differences abound.
Some of these differences are easy to characterize (e.g., subject-verb-object vs.~verb-subject-object, prepositions vs.~postpositions, adjective-noun vs.~noun-adjective), while others are subtle (e.g., number and positions of negation morphemes).  It is not at all clear how to translate descriptive facts about a language's syntax into features for a parser.

Consequently, training a language-universal parser on treebanks in multiple source languages requires caution.  While exposing the parser to a diverse set of syntactic patterns across many languages has the potential to improve its performance in each, dependency annotations in one language will, in some ways, contradict those in typologically different languages.

For instance, consider a context where the next word on the buffer is a noun, and the top word on the stack is an adjective, followed by a noun.
Treebanks of languages where postpositive adjectives are typical (e.g., French) will often teach the parser to predict \textsc{reduce-left}, while those of languages where prepositive adjectives are more typical (e.g., English) will teach the parser to predict \textsc{shift}.

Inspired by \newcite{naseem:12},
we address this problem by informing the parser about the input language it is currently parsing.  
Let $\mathbf{l}$ be the input vector representation of a particular language.
We consider three definitions for $\mathbf{l}$:\footnote{The files which contain these definitions are available at \url{https://github.com/clab/language-universal-parser/tree/master/typological_properties}.}
\begin{itemizesquish}
\item one-hot encoding of the language ID,
\item one-hot encoding of individual word-order properties,\footnote{The World Atlas of Language Structures (WALS; Dryer and Haspelmath, 2013) \nocite{dryer:13} is an online portal documenting typological properties of 2,679 languages (as of July 2015). We use the same set of WALS features used by \newcite{zhang:15}, namely 82A (order of subject and verb), 83A (order of object and verb), 85A (order of adposition and noun phrase), 86A (order of genitive and noun), and 87A (order of adjective and noun).} and 
\item averaged one-hot encoding of WALS typological properties (including word-order properties).\footnote{Some WALS features are not annotated for all languages. Therefore, we use the average value of all languages in the same genus. We rescale all values to be in the range $[-1, 1]$.}
\end{itemizesquish}

It is worth noting that the first definition (language ID) turns out to work best in our experiments.

We use a hidden layer with $\tanh$ nonlinearity to compute the language embedding $\mathbf{l'}$ as:
\begin{align}
\mathbf{l'} = \tanh(\mathbf{L l + L_{\text{bias}}}) \nonumber
\end{align}
where the matrix $\mathbf{L}$ and the vector $\mathbf{L_{\text{bias}}}$ are additional model parameters.
We modify the parsing architecture as follows:
\begin{itemizesquish}
\item include $\mathbf{l'}$ in the token representation (which feeds into the stack-LSTM modules of the buffer and the stack as described in \S\ref{sec:slstm}),
\item include $\mathbf{l'}$ in the action vector representation (which feeds into the stack-LSTM module that represents previous actions as described in \S\ref{sec:slstm}), and
\item redefine the parser state at time $t$ as $\mathbf{p}_t = \max\left\{ \boldsymbol{0}, \mathbf{W}[\mathbf{s}_t; \mathbf{b}_t; \mathbf{a}_t; \mathbf{l'}] + \mathbf{W}_{\text{bias}}\right\}$.
\end{itemizesquish}

Intuitively, the first two modifications allow the input language to influence the vector representation of the stack, the buffer and the list of actions.
The third modification allows the input language to influence the parser state which in turn is used to predict the next action.
In preliminary experiments, we found that adding the language embeddings at the token and action level is important.
We also experimented with computing more complex functions of ($\mathbf{s}_t, \mathbf{b}_t, \mathbf{a}_t, \mathbf{l'}$) to define the parser state, but they did not help.

\subsection{Fine-grained POS Tag Embeddings}
\label{sec:fine_grained}
\newcite{tiedemann:15} shows that omitting fine-grained POS tags significantly hurts the performance of a dependency parser.
However, those fine-grained POS tagsets are defined monolingually and are only available for a subset of the languages with universal dependency treebanks.

We extend the token representation to include a fine-grained POS embedding (in addition to the coarse POS embedding).
We stochastically dropout the fine-grained POS embedding for each token with 50\% probability \cite{srivastava:14} so that the parser can make use of fine-grained POS tags when available but stay reliable when the fine-grained POS tags are missing.

\subsection{Predicting POS Tags}
\label{sec:joint}
The model discussed thus far conditions on the POS tags of words in the input sentence.
However, gold POS tags may not be available in real applications (e.g., parsing the web).
Here, we describe two modifications to (i) model both POS tagging and dependency parsing, and (ii) increase the robustness of the parser to incorrect POS predictions.

\paragraph{Tagging model.}
Let $x_1, \ldots, x_n$, $y_1,\ldots, y_n$, $z_1, \ldots, z_{2n}$ be the sequence of words, POS tags, and parsing actions, respectively, for a sentence of length $n$.
We define the joint distribution of a POS tag sequence and parsing actions given a sequence of words as follows:
\begin{align}
p&(y_1,\ldots, y_n, z_1, \ldots,z_{2n} \mid x_1,\ldots,x_n) = \nonumber \\
&\prod_{i=1}^{n} p(y_i \mid x_1,\ldots,x_n) \nonumber \\
\times& \prod_{j=1}^{2n} p(z_j \mid x_1, \ldots, x_n, y_1, \ldots, y_n, z_1, \ldots, z_{j-1}) \nonumber
\end{align}
where $p(z_j \mid \ldots)$ is defined in Eq.~\ref{eq:action_prob}, and $p(y_i \mid x_1, \ldots, x_n)$ uses a bidirectional LSTM \cite{graves:13}.
\newcite{huang:15} show that the performance of a bidirectional LSTM POS tagger is on par with a conditional random field tagger.

We use slightly different token representations for tagging and parsing in the same model.
For \textit{tagging}, we construct the token representation by concatenating the embeddings of the word type (pretrained), the Brown cluster and the input language.
This token representation feeds into the bidirectional LSTM, followed by a softmax layer (at each position) which defines a categorical distribution over possible POS tags.
For \textit{parsing}, we construct the token representation by further concatenating the embeddings of predicted POS tags.
This token representation feeds into the stack-LSTM modules of the buffer and stack components of the transition-based parser.
This multi-task learning setup enables us to predict both POS tags and dependency trees in the same model.
We note that pretrained word embeddings, cluster embeddings and language embeddings are shared for tagging and parsing.

\paragraph{Block dropout.}
We use an independently developed variant of \textit{word dropout} \cite{iyyer:15}, which we call \textit{block dropout}.
The token representation used for parsing includes the embedding of predicted POS tags, which may be incorrect.
We introduce another modification which makes the parser more robust to incorrect POS tag predictions, by stochastically zeroing out the entire embedding of the POS tag.
While training the parser, we replace the POS embedding vector $\mathbf{e}$ with another vector (of the same dimensionality) stochastically computed as: $\mathbf{e'} = (1-b)/\mu \times \mathbf{e}$,
where $b \in \{0,1\}$ is a Bernoulli-distributed random variable with parameter $\mu$ which is initialized to 1.0 (i.e., always dropout, setting $b=1, \mathbf{e'} = 0$), and is dynamically updated to match the error rate of the POS tagger on the development set.
At test time, we never dropout the predicted POS embedding, i.e., $\mathbf{e'}=\mathbf{e}$.
Intuitively, this method extends the dropout method \cite{srivastava:14} to address structured noise in the input layer.

\begin{table*}[!]
\centering


\begin{tabular}{l|r|r|r|r|r|r|r||r}
LAS & \multicolumn{7}{c||}{target language} & average \\ \hline
                                                       & {de} & {en} & {es} & {fr} & {it} & {pt} & {sv} & {}   \\ \hline
monolingual                                   & \textbf{79.3} & \textbf{85.9} & 83.7 & 81.7 & 88.7 & 85.7 & 83.5 & 84.0 \\ \hline
\malopa                            & 70.4 & 69.3 & 72.4 & 71.1 & 78.0 & 74.1 & 65.4 & 71.5 \\
                   +lexical                   & 76.7 & 82.0 & 82.7 & 81.2 & 87.6 & 82.1 & 81.2 & 81.9 \\ 
\hspace{0.1cm} +language ID                   & 78.6 & 84.2 & 83.4 & \textbf{82.4} & \textbf{89.1} & 84.2 & 82.6 & 83.5 \\
\hspace{0.3cm} +fine-grained POS              & 78.9 & 85.4 & \textbf{84.3} & \textbf{82.4} & 89.0 & \textbf{86.2} & \textbf{84.5} & \textbf{84.3} \\ 
\end{tabular}

\caption{Dependency parsing:  labeled attachment scores (LAS) for monolingually-trained parsers and \malopa ~in the fully supervised scenario where $L^t = L^s$. Note that we use the universal dependencies verson 1.2 which only includes annotations for $\sim$13,000 English sentences, which explains the relatively low scores in English. When we instead use the universal dependency treebanks version 2.0 which includes annotations for $\sim$40,000 English sentences (originally from the English Penn Treebank), we achieve UAS score 93.0 and LAS score  91.5.
\label{tab:with}}
\end{table*}

\section{Experiments}

In this section, we evaluate the \malopa~approach in three data scenarios: when the target language has a large treebank (Table \ref{tab:with}), a small treebank (Table \ref{tab:small_treebank}) or no treebank (Table \ref{tab:without}).

\paragraph{Data.} For experiments where the target language has a large treebank, we use the standard data splits for German (de), English (en), Spanish (es), French (fr), Italian (it), Portuguese (pt) and Swedish (sv) in the latest release (version 1.2) of Universal Dependencies \cite{universal:v1_2}, and experiment with both gold and predicted POS tags.
For experiments where the target language has no treebank, we use the standard splits for these languages in the older universal dependency treebanks v2.0 \cite{mcdonald:13} and use gold POS tags, following the baselines \cite{zhang:15,guo:16}.
Table \ref{tab:treebanks} gives the number of sentences and words annotated for each language in both versions.
In a preprocessing step, we lowercase all tokens and remove multi-word annotations and language-specific dependency relations.
We use the same multilingual Brown clusters and multilingual embeddings of \newcite{guo:16}, kindly provided by the authors.

\paragraph{Optimization.} We follow \newcite{dyer:15} in parameter initialization and optimization.\footnote{We use stochastic gradient updates with an initial learning rate of $\eta_0=0.1$ in epoch \#0, update the learning rate in following epochs as $\eta_t=\eta_0/(1+ 0.1 t)$. We clip the $\ell_2$ norm of the gradient to avoid ``exploding'' gradients. Unlabeled attachment score (UAS) on the development set determines early stopping. Parameters are initialized with uniform samples in $\pm \sqrt{6/(r+c)}$ where $r$ and $c$ are the sizes of the previous and following layer in the nueral network \cite{glorot:10}. The standard deviations of the labeled attachment score (LAS) due to random initialization in individual target languages are 0.36 (de), 0.40 (en), 0.37 (es), 0.46 (fr), 0.47 (it), 0.41 (pt) and 0.24 (sv). The standard deviation of the average LAS scores across languages is 0.17.}
However, when training the parser on multiple languages in \malopa, instead of updating the parameters with the gradient of individual sentences, we use mini-batch updates which include one sentence sampled uniformly (without replacement) from each language's treebank, until all sentences in the smallest treebank are used (which concludes an epoch). 
We repeat the same process in following epochs.
We found this to help prevent one source language with a larger treebank (e.g., German) from dominating parameter updates at the expense of other source languages with a smaller treebank (e.g., Swedish).

\subsection{Target Languages with a Treebank ($L^t = L^s$)}
\label{sec:with}
Here, we evaluate our \malopa~parser when the target language has a treebank.

\paragraph{Baseline.} For each target language, the strong baseline we use is a monolingually-trained S-LSTM parser with a token representation which concatenates: pretrained word embeddings (50 dimensions),\footnote{These embeddings are treated as fixed inputs to the parser, and are not optimized towards the parsing objective. We use the same embeddings used in \newcite{guo:16}.} learned word embeddings (50 dimensions), coarse (universal) POS tag embeddings (12 dimensions), fine-grained (language-specific, when available) POS tag embeddings (12 dimensions), and embeddings of Brown clusters (12 dimensions), and uses a two-layer S-LSTM for each of the stack, the buffer and the list of actions.
We independently train one baseline parser for each target language, and share no model parameters.
This baseline, denoted `monolingual' in Tables \ref{tab:with} and \ref{tab:small_treebank}, achieves UAS score 93.0 and LAS score 91.5 when trained on the English Penn Treebank, which is comparable to \newcite{dyer:15}.

\begin{table*}[!]
\centering
\scalebox{0.92}{
\begin{tabular}{l|cc|c|cc|ccccccc}
Recall \%                                              & left & right & root & short & long & nsubj* &  dobj    & conj        & *comp      & case & *mod  \\ \hline
{monolingual}                                   & \textbf{89.9} & 95.2  & 86.4 & 92.9  & 81.1 & \textbf{77.3}    &  \textbf{75.5}   & 66.0        & 45.6       & \textbf{93.3} & \textbf{77.0}  \\ \hline
\malopa                                                & 85.4 & 93.3  & 80.2 & 91.2  & 73.3 & 57.3    &  62.7   & 64.2        & 34.0       & 90.7 & 69.6  \\
{           +lexical     }                      & \textbf{89.9} & 93.8  & 84.5 & 92.6  & 78.6 & 73.3    &  73.4   & 66.9        & 35.3       & 91.6 & 75.3  \\
{\hspace{0.1cm} +language ID}                   & 89.1 & 94.7  & 86.6 & 93.2  & 81.4 & 74.7    &  73.0   & \textbf{71.2}        & \textbf{48.2}       & 92.8 & 76.3  \\
{\hspace{0.3cm} +fine-grained POS}              & 89.5 & \textbf{95.7}  & \textbf{87.8} & \textbf{93.6}  & \textbf{82.0} & 74.7    &  74.9   & 69.7        & 46.0       & \textbf{93.3} & 76.3  \\
\end{tabular}

}

\caption{Recall of some classes of dependency attachments/relations in German.
\label{tab:analysis}}
\end{table*}

\begin{table*}[!]
\centering


\begin{tabular}{lcc||r|r|r|r|r|r|r||r}
LAS & & & \multicolumn{7}{c||}{target language} & average \\ \hline
    & language ID & coarse POS                     & {de} & {en} & {es} & {fr} & {it} & {pt} & {sv} & {}   \\ \hline
    & gold        & gold                           & 78.6 & 84.2 & 83.4 & 82.4 & 89.1 & 84.2 & 82.6 & 83.5 \\
    & predicted   & gold                           & 78.5 & 80.2 & 83.4 & 82.1 & 88.9 & 83.9 & 82.5 & 82.7 \\
    & gold        & predicted                      & 71.2 & 79.9 & 80.5 & 78.5 & 85.0 & 78.4 & 75.5 & 78.4 \\
    & predicted   & predicted                      & 70.8 & 74.1 & 80.5 & 78.2 & 84.7 & 77.1 & 75.5 & 77.2
\end{tabular}

\caption{Effect of automatically predicting language ID and POS tags with \malopa~ on LAS scores.\label{tab:langid_pos}}
\end{table*}

\paragraph{\malopa.}
We train  \malopa~ on the concantenation of training sections of all seven languages.
To balance the development set, we only concatenate the first 300 sentences of each language's development section.

\paragraph{Token representations.}
The first \malopa~parser we evaluate uses only coarse POS embeddings to construct the token representation.\footnote{We use the same number of dimensions for the coarse POS embeddings as in the monolingual baselines. The same applies to all other types of embeddings used in \malopa.}
As shown in Table \ref{tab:with}, this parser consistently underperforms the monolingual baselines, with a gap of 12.5 LAS points on average.

Augmenting the token representation with lexical embeddings to the token representation (both multilingual word clusters and pretrained multilingual word embeddings, as described in \S\ref{sec:embeddings}) substantially improves the performance of \malopa, recovering 83\% of the gap in average performance.

We experimented with three ways to include language information in the token representation, namely: `language ID', `word order' and `full typology' (see \S\ref{sec:lang} for details), and found all three to improve the performance of \malopa~ giving LAS scores 83.5, 83.2 and 82.5, respectively.
It is noteworthy that the model benefits more from language ID than from typological properties.
Using `language ID,' we recover another 12\% of the original gap.

Finally, the best configuration of \malopa~adds fine-grained POS embeddings to the token representation.\footnote{Fine-grained POS tags were only available for English, Italian, Portuguese and Swedish. Other languages reuse the coarse POS tags as fine-grained tags instead of padding the extra dimensions in the token representation with zeros.}
Surprisingly, adding fine-grained POS embeddings improves the performance even for some languages where fine-grained POS tags are not available (e.g., Spanish).
This parser outperforms the monolingual baseline in five out of seven target languages, and wins on average by 0.3 LAS points.
We emphasize that this model is only trained once on all languages, and the \textit{same model} is used to parse the test set of each language, which simplifies the distribution or deployment of multilingual parsing software.

\paragraph{Qualitative analysis.}
To gain a better understanding of the model behavior, we analyze certain classes of dependency attachments/relations in German, which has notably flexible word order, in Table \ref{tab:analysis}.
We consider the recall of left attachments (where the head word precedes the dependent word in the sentence), right attachments, root attachments, short-attachments (with distance $=1$), long-attachments (with distance $>6$), as well as the following relation groups: nsubj* (nominal subjects: \texttt{nsubj}, \texttt{nsubjpass}), dobj (direct object: \texttt{dobj}), conj (conjunct: \texttt{conj}), *comp (clausal complements: \texttt{ccomp}, \texttt{xcomp}), case (clitics and adpositions: \texttt{case}), *mod (modifiers of a noun: \texttt{nmod}, \texttt{nummod}, \texttt{amod}, \texttt{appos}), neg (negation modifier: \texttt{neg}).\footnote{For each group, we report recall of both the attachment and relation weighted by the number of instances in the gold annotation. A detailed description of each relation can be found at \url{http://universaldependencies.org/u/dep/index.html}}

\paragraph{Findings.} 
We found that each of the three improvements (lexical embeddings, language embeddings and fine-grained POS embeddings) tends to improve recall for most classes.
\malopa~ underperforms (compared to the monolingual baseline) in some classes: nominal subjects, direct objects and modifiers of a noun.
Nevertheless, \malopa~ outperforms the baseline in some important classes such as: root, long attachments and conjunctions.

\paragraph{Predicting language IDs and POS tags.}
In Table~\ref{tab:with}, we assume that both gold language ID of the input language and gold POS tags are given at test time.
However, this assumption is not realistic in practical applications.
Here, we quantify the degradation in parsing accuracy when language ID and POS tags are only given at training time, but must be predicted at test time. 
We do not use fine-grained POS tags in these experiments because some languages use a very large fine-grained POS tag set (e.g., 866 unique tags in Portuguese).

In order to predict language ID, we use the \texttt{langid.py} library \cite{lui:12}\footnote{\url{https://github.com/saffsd/langid.py}} and classify individual sentences in the test sets to one of the seven languages of interest, using the default models included in the library.
The macro average language ID prediction accuracy on the test set across sentences is 94.7\%.
In order to predict POS tags, we use the model described in \S\ref{sec:joint} with both input and hidden LSTM dimensions of 60, and with block dropout.
The macro average accuracy of the POS tagger is 93.3\%.
Table \ref{tab:langid_pos} summarizes the four configurations: \{gold language ID, predicted language ID\} $\times$ \{gold POS tags, predicted POS tags\}.
The performance of the parser suffers mildly (--0.8 LAS points) when using predicted language IDs, but more (--5.1 LAS points) when using predicted POS tags.
As an alternative approach to predicting POS tags, we trained the Stanford POS tagger, for each target language, on the coarse POS tag annotations in the training section of the universal dependency treebanks,\footnote{We used version 3.6.0 of the Stanford POS tagger, with the following pre-packaged configuration files: german-fast-caseless.tagger.props (de), english-caseless-left3words-distsim.tagger.props (en), spanish.tagger.props (es), french.tagger.props (fr). We reused french.tagger.props for (it, pt, sv).} then replaced the gold POS tags in the test set of each language with predictions of the monolingual tagger.
The resulting degradation in parsing performance between gold vs.~predicted POS tags is --6.0 LAS points (on average, compared to a degradation of --5.1 LAS points in Table \ref{tab:langid_pos}).
The disparity in parsing results with gold vs.~predicted POS tags is an important open problem, and has been previously discussed by \newcite{tiedemann:15}.

The predicted POS results in Table \ref{tab:langid_pos} use block dropout.
Without using block dropout, we lose an extra 0.2 LAS points in both configurations using predicted POS tags.

\paragraph{Different multilingual embeddings.}

\begin{table}[!]
\centering
\begin{tabular}{r|r|r}
multilingual embeddings           & UAS & LAS   \\ \hline
multiCluster           & 87.7  & 84.1   \\
multiCCA               & 87.8  & 84.4   \\
robust projection      & 87.8  & 84.2
\end{tabular}

\caption{Effect of multilingual embedding estimation method on the multilingual parsing with \malopa.
UAS and LAS scores are macro-averaged across seven target languages.
\label{tab:multilingual_embeddings}}
\end{table}

Several methods have been proposed for pretraining multilingual word embeddings.
We compare three of them:
\begin{itemize}
\item multiCCA \cite{ammar:16} uses a linear operator to project pretrained monolingual embeddings in each language (except English) to the vector space of pretrained English word embeddings.
\item multiCluster \cite{ammar:16} uses the same embedding for translationally-equivalent words in different languages.
\item robust projection \cite{guo:15} first pretrains monolingual English word embeddings, then defines the embedding of a non-English word as the weighted average embedding of English words aligned to the non-English words (in a parallel corpus). The embedding of a non-English word which is not aligned to any English words is defined as the average embedding of words with a unit edit distance in the same language (e.g., `playz' is the average of `plays' and `play').\footnote{Our implementation of this method can be found at \url{https://github.com/gmulcaire/average-embeddings}.}
\end{itemize}
All embeddings are trained on the same data and use the same number of dimensions (100).\footnote{We share the embedding files at \url{https://github.com/clab/language-universal-parser/tree/master/pretrained_embeddings}.}
Table \ref{tab:multilingual_embeddings} illustrates that the three methods perform similarly on this task.
Aside from Table \ref{tab:multilingual_embeddings}, in this paper, we exclusively use the robust projection multilingual embeddings trained in \newcite{guo:16}.\footnote{The embeddings were kindly provided by the authors of \newcite{guo:16} at \url{https://drive.google.com/file/d/0B1z04ix6jD_DY3lMN2Ntdy02NFU/view}}
The ``robust projection'' result in Table \ref{tab:multilingual_embeddings} (which uses 100 dimensions) is comparable to the last row in Table \ref{tab:with} (which uses 50 dimensions).

\begin{table}[!]
\centering
\begin{tabular}{l|r|r|r|r|r}
LAS & \multicolumn{5}{c}{target language} \\ \hline
                   & {de} & {es} & {fr} & {it} & {sv}   \\ \hline
monolingual        & 58.0 & 64.7 & 63.0 & 68.7 & 57.6 \\ \hline
Duong et al.       & 61.8 & \textbf{70.5} & 67.2 & 71.3 & 62.5  \\
\malopa          & \textbf{63.4} & \textbf{70.5} & \textbf{69.1} & \textbf{74.1} & \textbf{63.4} 
\end{tabular}

\caption{Small (3,000 token) target treebank setting:  language-universal dependency parser performance.
\label{tab:small_treebank}}
\end{table}

\begin{table*}[!]
\centering


\begin{tabular}{l|r|r|r|r|r|r||r}
LAS & \multicolumn{6}{c||}{target language} & average \\ \hline
                   & {de} & {es} & {fr} & {it} & {pt} & {sv} & {}   \\ \hline
\newcite{zhang:15}     & 54.1 & 68.3 & 68.8 & 69.4 & 72.5 & 62.5 & 65.9 \\
\newcite{guo:16} & 55.9 & 73.0 & 71.0 & 71.2 & 78.6 & 69.5 & 69.3 \\ \hline
\malopa    & 57.1 & 74.6 & 73.9 & 72.5 & 77.0 & 68.1 & 70.5
\end{tabular}

\caption{Dependency parsing:  labeled attachment scores (LAS) for multi-source transfer parsers  in the simulated low-resource scenario where $L^t \cap L^s = \emptyset$. 
\label{tab:without}}
\end{table*}

\paragraph{Small target treebank.}
\newcite{duong:15} considered a setup where the target language has a small treebank of $\sim$3,000 tokens, and the source language (English) has a large treebank of $\sim$205,000 tokens.
The parser proposed in \newcite{duong:15} is a neural network parser based on \newcite{chen:14}, which shares most of the parameters between English and the target language, and uses an $\ell_2$ regularizer to tie the lexical embeddings of translationally-equivalent words.
While not the primary focus of this paper,\footnote{The setup cost involved in recruiting linguists, developing and revising annotation guidelines to annotate a new language ought to be higher than the cost of annotating 3,000 tokens. After investing much resources in a language, we believe it is unrealistic to stop the annotation effort after only 3,000 tokens.}
we compare our proposed method to that of \newcite{duong:15} on five target languages for which multilingual Brown clusters are available from \newcite{guo:16}.
For each target language, we train the parser on the English training data in the UD version 1.0 corpus \cite{universal:v1_0} and a small treebank in the target language.\footnote{We thank Long Duong for sharing the processed, subsampled training corpora in each target language at \url{https://github.com/longdt219/universal_dependency_parser/tree/master/data/universal-dep/universal-dependencies-1.0}.}
Following \newcite{duong:15}, in this setup, we only use gold coarse POS tags, we do not use any development data in the target languages (we use the English development set instead), and we subsample the English training data in each epoch to the same number of sentences in the target language.
We use the same hyperparameters specified before for the single \malopa~parser and each of the monolingual baselines.
Table~\ref{tab:small_treebank} shows that our method outperforms \newcite{duong:15} by 1.4 LAS points on average.
Our method consistently outperforms the monolingual baselines in this setup, with an average improvement of 5.7 absolute LAS points.

\subsection{Target Languages without a Treebank ($L^t \cap L^s = \emptyset$)}

\newcite{mcdonald:11} established that, when no treebank annotations are available in the target language, training on multiple source languages outperforms training on one (i.e., multi-source model transfer outperforms single-source model transfer).
In this section, we evaluate the performance of our parser in this setup.
We use two strong baseline multi-source model transfer parsers with no supervision in the target language:
\begin{itemizesquish}
\item \newcite{zhang:15} is a graph-based arc-factored parsing model with a tensor-based scoring function. It takes typological properties of a language as input.
We compare to the best reported configuration (i.e., the column titled ``OURS'' in Table 5 of Zhang and Barzilay, 2015).
\item \newcite{guo:16} is a transition-based neural-network parsing model based on \newcite{chen:14}. It uses a multilingual embeddings and Brown clusters as lexical features.
We compare to the best reported configuration (i.e., the column titled ``MULTI-PROJ'' in Table 1 of Guo et al., 2016).
\end{itemizesquish}
Following \newcite{guo:16}, for each target language, we train the parser on six other languages in the Google universal dependency treebanks version 2.0\footnote{\url{https://github.com/ryanmcd/uni-dep-tb/}} (de, en, es, fr, it, pt, sv, excluding whichever is the target language), and we use gold coarse POS tags.
Our parser uses the same word embeddings and word clusters used in \newcite{guo:16}, and does not use any typology information.\footnote{In preliminary experiments, we found language embeddings to hurt the performance of the parser for target languages without a treebank.}

The results in Table \ref{tab:without} show that, on average, our parser outperforms both baselines by more than 1 point in LAS, and gives the best LAS results in four (out of six) languages.

\section{Related Work}
\label{sec:related}
Our work builds on the model transfer approach, which was pioneered by \newcite{zeman:08} who trained a parser on a source language treebank then applied it to parse sentences in a target language.
\newcite{cohen:11} and \newcite{mcdonald:11} trained unlexicalized parsers on treebanks of multiple source languages and applied the parser to different languages.
\newcite{naseem:12}, \newcite{tackstrom:13}, and \newcite{zhang:15} used language typology to improve model transfer.
To add lexical information, \newcite{tackstrom:12} used multilingual word clusters, while \newcite{xiao:14}, \newcite{guo:15}, \newcite{sogaard:15} and \newcite{guo:16} used multilingual word embeddings.
\newcite{duong:15} used a neural network based model, sharing most of the parameters between two languages, and used an $\ell_2$ regularizer to tie the lexical embeddings of translationally-equivalent words.
We incorporate these ideas in our framework, while proposing a novel neural architecture for embedding language typology (see \S\ref{sec:lang}), and use a variant of word dropout \cite{iyyer:15} for consuming noisy structured inputs.
We also show how to replace an array of monolingually trained parsers with one multilingually-trained parser without sacrificing accuracy, which is related to \newcite{vilares:16}.

Neural network parsing models which preceded \newcite{dyer:15} include \newcite{henderson:03}, \newcite{titov:07}, \newcite{henderson:10} and \newcite{chen:14}.
Related to lexical features in cross-lingual parsing is \newcite{durrett:12} who defined lexico-syntactic features based on bilingual lexicons.
Other related work include \newcite{ostling:15}, which may be used to induce more useful typological properties to inform multilingual parsing.

Another popular approach for cross-lingual supervision is to project annotations from the source language to the target language via a parallel corpus \cite{yarowsky:01,hwa:05} or via automatically-translated sentences \cite{tiedemann:14b}. 
\newcite{ma:14} used entropy regularization to learn from both parallel data (with projected annotations) and unlabeled data in the target language.
\newcite{rasooli:15} trained an array of target-language parsers on fully annotated trees, by iteratively decoding sentences in the target language with incomplete annotations. 
One research direction worth pursuing is to find synergies between the model transfer approach and annotation projection approach.

\section{Conclusion}
We presented \malopa, a single parser trained on a multilingual set of treebanks.
We showed that this parser, equipped with language embeddings and fine-grained POS embeddings, on average outperforms monolingually-trained parsers for target languages with a treebank. This pattern of results is quite encouraging. Although languages may share underlying syntactic properties, individual parsing models must behave quite differently, and our model allows this while sharing parameters across languages. The value of this sharing is more pronounced in scenarios where the target language's training treebank is small or non-existent, where our parser outperforms previous cross-lingual multi-source model transfer methods.

\section*{Acknowledgments}
Waleed Ammar is supported by the Google fellowship in natural language processing.
Miguel Ballesteros is supported by the European Commission under the contract numbers FP7-ICT-610411 (project MULTISENSOR) and H2020-RIA-645012 (project KRISTINA).
Part of this material is based upon work supported by a subcontract with Raytheon BBN Technologies Corp. under DARPA Prime Contract  No.~HR0011-15-C-0013, and part of this research was supported by a Google research award to Noah Smith.
We thank Jiang Guo for sharing the multilingual word embeddings and multilingual word clusters.
We thank Lori Levin, Ryan McDonald, J\"{o}rg Tiedemann, Yulia Tsvetkov, and Yuan Zhang for helpful discussions.
Last but not least, we thank the anonymous TACL reviewers for their valuable feedback.

\bibliographystyle{naaclhlt2015}
\bibliography{biblio}

\begin{thebibliography}{}

\bibitem[\protect\citename{Agi{\'c} \bgroup et al.\egroup
  }2015]{universal:v1_1}
{\v Z}eljko Agi{\'c}, Maria~Jesus Aranzabe, Aitziber Atutxa, Cristina Bosco,
  Jinho Choi, Marie-Catherine de~Marneffe, Timothy Dozat, Rich{\'a}rd Farkas,
  Jennifer Foster, Filip Ginter, Iakes Goenaga, Koldo Gojenola, Yoav Goldberg,
  Jan Haji{\v c}, Anders~Tr{\ae}rup Johannsen, Jenna Kanerva, Juha Kuokkala,
  Veronika Laippala, Alessandro Lenci, Krister Lind{\'e}n, Nikola Ljube{\v
  s}i{\'c}, Teresa Lynn, Christopher Manning, H{\'e}ctor~Alonso Mart{\'i}nez,
  Ryan {McDonald}, Anna Missil{\"a}, Simonetta Montemagni, Joakim Nivre, Hanna
  Nurmi, Petya Osenova, Slav Petrov, Jussi Piitulainen, Barbara Plank, Prokopis
  Prokopidis, Sampo Pyysalo, Wolfgang Seeker, Mojgan Seraji, Natalia Silveira,
  Maria Simi, Kiril Simov, Aaron Smith, Reut Tsarfaty, Veronika Vincze, and
  Daniel Zeman.
\newblock 2015.
\newblock Universal dependencies 1.1.
\newblock {LINDAT}/{CLARIN} digital library at Institute of Formal and Applied
  Linguistics, Charles University in Prague.

\bibitem[\protect\citename{Ammar \bgroup et al.\egroup }2016]{ammar:16}
Waleed Ammar, George Mulcaire, Yulia Tsvetkov, Guillaume Lample, Chris Dyer,
  and Noah~A. Smith.
\newblock 2016.
\newblock Massively multilingual word embeddings.
\newblock arXiv:1602.01925v2.

\bibitem[\protect\citename{Barman \bgroup et al.\egroup }2014]{barman:14}
Utsab Barman, Amitava Das, Joachim Wagner, and Jennifer Foster.
\newblock 2014.
\newblock Code mixing: A challenge for language identification in the language
  of social media.
\newblock In {\em EMNLP Workshop on Computational Approaches to Code
  Switching}.

\bibitem[\protect\citename{Bender}2011]{bender:11}
Emily~M. Bender.
\newblock 2011.
\newblock On achieving and evaluating language-independence in {NLP}.
\newblock {\em Linguistic Issues in Language Technology}, 6(3):1--26.

\bibitem[\protect\citename{Buchholz and Marsi}2006]{buchholz:06}
Sabine Buchholz and Erwin Marsi.
\newblock 2006.
\newblock {CoNLL-X} shared task on multilingual dependency parsing.
\newblock In {\em Proc. of CoNLL}.

\bibitem[\protect\citename{Chen and Manning}2014]{chen:14}
Danqi Chen and Christopher Manning.
\newblock 2014.
\newblock A fast and accurate dependency parser using neural networks.
\newblock In {\em Proc. of EMNLP}.

\bibitem[\protect\citename{Cohen \bgroup et al.\egroup }2011]{cohen:11}
Shay~B. Cohen, Dipanjan Das, and Noah~A. Smith.
\newblock 2011.
\newblock Unsupervised structure prediction with non-parallel multilingual
  guidance.
\newblock In {\em Proc. of EMNLP}.

\bibitem[\protect\citename{Dryer and Haspelmath}2013]{dryer:13}
Matthew~S. Dryer and Martin Haspelmath, editors.
\newblock 2013.
\newblock {\em WALS Online}.
\newblock Max Planck Institute for Evolutionary Anthropology, Leipzig.

\bibitem[\protect\citename{Duong \bgroup et al.\egroup }2015a]{duong:15b}
Long Duong, Trevor Cohn, Steven Bird, and Paul Cook.
\newblock 2015a.
\newblock Low resource dependency parsing: Cross-lingual parameter sharing in a
  neural network parser.
\newblock In {\em Proc. of ACL-IJCNLP}.

\bibitem[\protect\citename{Duong \bgroup et al.\egroup }2015b]{duong:15}
Long Duong, Trevor Cohn, Steven Bird, and Paul Cook.
\newblock 2015b.
\newblock A neural network model for low-resource universal dependency parsing.
\newblock In {\em Proc. of EMNLP}.

\bibitem[\protect\citename{Durrett \bgroup et al.\egroup }2012]{durrett:12}
Greg Durrett, Adam Pauls, and Dan Klein.
\newblock 2012.
\newblock Syntactic transfer using a bilingual lexicon.
\newblock In {\em Proc. of EMNLP}.

\bibitem[\protect\citename{Dyer \bgroup et al.\egroup }2015]{dyer:15}
Chris Dyer, Miguel Ballesteros, Wang Ling, Austin Matthews, and Noah~A. Smith.
\newblock 2015.
\newblock Transition-based dependency parsing with stack long short-term
  memory.
\newblock In {\em Proc. of ACL}.

\bibitem[\protect\citename{Gardner \bgroup et al.\egroup }2015]{gardner:15}
Matt Gardner, Kejun Huang, Evangelos Papalexakis, Xiao Fu, Partha Talukdar,
  Christos Faloutsos, Nicholas Sidiropoulos, and Tom Mitchell.
\newblock 2015.
\newblock Translation invariant word embeddings.
\newblock In {\em Proc. of EMNLP}.

\bibitem[\protect\citename{Glorot and Bengio}2010]{glorot:10}
Xavier Glorot and Yoshua Bengio.
\newblock 2010.
\newblock Understanding the difficulty of training deep feedforward neural
  networks.
\newblock In {\em Proc. of AISTATS}.

\bibitem[\protect\citename{Graves \bgroup et al.\egroup }2013]{graves:13}
Alan Graves, Abdel-rahman Mohamed, and Geoffrey Hinton.
\newblock 2013.
\newblock Speech recognition with deep recurrent neural networks.
\newblock In {\em Proc. of ICASSP}.

\bibitem[\protect\citename{Guo \bgroup et al.\egroup }2015]{guo:15}
Jiang Guo, Wanxiang Che, David Yarowsky, Haifeng Wang, and Ting Liu.
\newblock 2015.
\newblock Cross-lingual dependency parsing based on distributed
  representations.
\newblock In {\em Proc. of ACL}.

\bibitem[\protect\citename{Guo \bgroup et al.\egroup }2016]{guo:16}
Jiang Guo, Wanxiang Che, David Yarowsky, Haifeng Wang, and Ting Liu.
\newblock 2016.
\newblock A representation learning framework for multi-source transfer
  parsing.
\newblock In {\em Proc. of AAAI}.

\bibitem[\protect\citename{Heid and Raab}1989]{heid:89}
Ulrich Heid and Sybille Raab.
\newblock 1989.
\newblock Collocations in multilingual generation.
\newblock In {\em Proc. of EACL}.

\bibitem[\protect\citename{Henderson and Titov}2010]{henderson:10}
James Henderson and Ivan Titov.
\newblock 2010.
\newblock Incremental sigmoid belief networks for grammar learning.
\newblock {\em Journal of Machine Learning Research}, 11:3541--3570.

\bibitem[\protect\citename{Henderson}2003]{henderson:03}
James Henderson.
\newblock 2003.
\newblock Inducing history representations for broad coverage statistical
  parsing.
\newblock In {\em Proc. of NAACL-HLT}.

\bibitem[\protect\citename{Huang \bgroup et al.\egroup }2015]{huang:15}
Zhiheng Huang, Wei Xu, and Kai Yu.
\newblock 2015.
\newblock Bidirectional {LSTM-CRF} models for sequence tagging.
\newblock arXiv:1508.01991.

\bibitem[\protect\citename{Hwa \bgroup et al.\egroup }2005]{hwa:05}
Rebecca Hwa, Philip Resnik, Amy Weinberg, Clara Cabezas, and Okan Kolak.
\newblock 2005.
\newblock Bootstrapping parsers via syntactic projection across parallel texts.
\newblock {\em Natural Language Engineering}, 11(03):311--325.

\bibitem[\protect\citename{Iyyer \bgroup et al.\egroup }2015]{iyyer:15}
Mohit Iyyer, Varun Manjunatha, Jordan~L. Boyd-Graber, and Hal Daum{\'e}.
\newblock 2015.
\newblock Deep unordered composition rivals syntactic methods for text
  classification.
\newblock In {\em Proc. of ACL}.

\bibitem[\protect\citename{Lui and Baldwin}2012]{lui:12}
Marco Lui and Timothy Baldwin.
\newblock 2012.
\newblock {l}angid.py: An off-the-shelf language identification tool.
\newblock In {\em Proc. of ACL}.

\bibitem[\protect\citename{Ma and Xia}2014]{ma:14}
Xuezhe Ma and Fei Xia.
\newblock 2014.
\newblock Unsupervised dependency parsing with transferring distribution via
  parallel guidance and entropy regularization.
\newblock In {\em Proc. of ACL}.

\bibitem[\protect\citename{McDonald \bgroup et al.\egroup }2011]{mcdonald:11}
Ryan McDonald, Slav Petrov, and Keith Hall.
\newblock 2011.
\newblock Multi-source transfer of delexicalized dependency parsers.
\newblock In {\em Proc. of EMNLP}.

\bibitem[\protect\citename{McDonald \bgroup et al.\egroup }2013]{mcdonald:13}
Ryan McDonald, Joakim Nivre, Yvonne Quirmbach-Brundage, Yoav Goldberg, Dipanjan
  Das, Kuzman Ganchev, Keith Hall, Slav Petrov, Hao Zhang, Oscar
  T\"{a}ckstr\"{o}m, Claudia Bedini, N\'{u}ria Bertomeu~Castell\'{o}, and
  Jungmee Lee.
\newblock 2013.
\newblock Universal dependency annotation for multilingual parsing.
\newblock In {\em Proc. of ACL}.

\bibitem[\protect\citename{Mikolov \bgroup et al.\egroup }2013]{mikolov:13}
Tomas Mikolov, Kai Chen, Greg Corrado, and Jeffrey Dean.
\newblock 2013.
\newblock Efficient estimation of word representations in vector space.
\newblock In {\em Proc. of ICLR}.

\bibitem[\protect\citename{Naseem \bgroup et al.\egroup }2012]{naseem:12}
Tahira Naseem, Regina Barzilay, and Amir Globerson.
\newblock 2012.
\newblock Selective sharing for multilingual dependency parsing.
\newblock In {\em Proc. of ACL}.

\bibitem[\protect\citename{Nivre and Nilsson}2005]{nivre:05}
Joakim Nivre and Jens Nilsson.
\newblock 2005.
\newblock Pseudo-projective dependency parsing.
\newblock In {\em Proc. of ACL}.

\bibitem[\protect\citename{Nivre \bgroup et al.\egroup }2007]{nivre:07}
Joakim Nivre, Johan Hall, Sandra Kubler, Ryan McDonald, Jens Nilsson, Sebastian
  Riedel, and Deniz Yuret.
\newblock 2007.
\newblock The {CoNLL} 2007 shared task on dependency parsing.
\newblock In {\em Proc. of CoNLL}.

\bibitem[\protect\citename{Nivre \bgroup et al.\egroup }2015a]{universal:v1_2}
Joakim Nivre, {\v Z}eljko Agi{\'c}, Maria~Jesus Aranzabe, Masayuki Asahara,
  Aitziber Atutxa, Miguel Ballesteros, John Bauer, Kepa Bengoetxea, Riyaz~Ahmad
  Bhat, Cristina Bosco, Sam Bowman, Giuseppe G.~A. Celano, Miriam Connor,
  Marie-Catherine de~Marneffe, Arantza Diaz~de Ilarraza, Kaja Dobrovoljc,
  Timothy Dozat, Toma{\v z} Erjavec, Rich{\'a}rd Farkas, Jennifer Foster,
  Daniel Galbraith, Filip Ginter, Iakes Goenaga, Koldo Gojenola, Yoav Goldberg,
  Berta Gonzales, Bruno Guillaume, Jan Haji{\v c}, Dag Haug, Radu Ion, Elena
  Irimia, Anders Johannsen, Hiroshi Kanayama, Jenna Kanerva, Simon Krek,
  Veronika Laippala, Alessandro Lenci, Nikola Ljube{\v s}i{\'c}, Teresa Lynn,
  Christopher Manning, Cătălina Mărănduc, David Mare{\v c}ek, H{\'e}ctor
  Mart{\'i}nez~Alonso, Jan Ma{\v s}ek, Yuji Matsumoto, Ryan {McDonald}, Anna
  Missil{\"a}, Verginica Mititelu, Yusuke Miyao, Simonetta Montemagni, Shunsuke
  Mori, Hanna Nurmi, Petya Osenova, Lilja {\O}vrelid, Elena Pascual, Marco
  Passarotti, Cenel-Augusto Perez, Slav Petrov, Jussi Piitulainen, Barbara
  Plank, Martin Popel, Prokopis Prokopidis, Sampo Pyysalo, Loganathan Ramasamy,
  Rudolf Rosa, Shadi Saleh, Sebastian Schuster, Wolfgang Seeker, Mojgan Seraji,
  Natalia Silveira, Maria Simi, Radu Simionescu, Katalin Simk{\'o}, Kiril
  Simov, Aaron Smith, Jan {\v S}t{\v e}p{\'a}nek, Alane Suhr, Zsolt
  Sz{\'a}nt{\'o}, Takaaki Tanaka, Reut Tsarfaty, Sumire Uematsu, Larraitz Uria,
  Viktor Varga, Veronika Vincze, Zden{\v e}k {\v Z}abokrtsk{\'y}, Daniel Zeman,
  and Hanzhi Zhu.
\newblock 2015a.
\newblock Universal dependencies 1.2.
\newblock {LINDAT}/{CLARIN} digital library at Institute of Formal and Applied
  Linguistics, Charles University in Prague.

\bibitem[\protect\citename{Nivre \bgroup et al.\egroup }2015b]{universal:v1_0}
Joakim Nivre, Cristina Bosco, Jinho Choi, Marie-Catherine de~Marneffe, Timothy
  Dozat, Rich{\'a}rd Farkas, Jennifer Foster, Filip Ginter, Yoav Goldberg, Jan
  Haji{\v c}, Jenna Kanerva, Veronika Laippala, Alessandro Lenci, Teresa Lynn,
  Christopher Manning, Ryan {McDonald}, Anna Missil{\"a}, Simonetta Montemagni,
  Slav Petrov, Sampo Pyysalo, Natalia Silveira, Maria Simi, Aaron Smith, Reut
  Tsarfaty, Veronika Vincze, and Daniel Zeman.
\newblock 2015b.
\newblock Universal dependencies 1.0.
\newblock {LINDAT}/{CLARIN} digital library at Institute of Formal and Applied
  Linguistics, Charles University in Prague.

\bibitem[\protect\citename{Nivre}2004]{nivre:04}
Joakim Nivre.
\newblock 2004.
\newblock Incrementality in deterministic dependency parsing.
\newblock In {\em Proceedings of the Workshop on Incremental Parsing: Bringing
  Engineering and Cognition Together}.

\bibitem[\protect\citename{{\"O}stling}2015]{ostling:15}
Robert {\"O}stling.
\newblock 2015.
\newblock Word order typology through multilingual word alignment.
\newblock In {\em Proc. of ACL-IJCNLP}.

\bibitem[\protect\citename{Petrov \bgroup et al.\egroup }2012]{petrov:12}
Slav Petrov, Dipanjan Das, and Ryan McDonald.
\newblock 2012.
\newblock A universal part-of-speech tagset.
\newblock In {\em Proc. of LREC}.

\bibitem[\protect\citename{Rasooli and Collins}2015]{rasooli:15}
Mohammad~Sadegh Rasooli and Michael Collins.
\newblock 2015.
\newblock Density-driven cross-lingual transfer of dependency parsers.
\newblock In {\em Proc. of EMNLP}.

\bibitem[\protect\citename{R{\"o}sner}1988]{rosner:88}
Deitmar R{\"o}sner.
\newblock 1988.
\newblock The generation system of the semsyn project: Towards a
  task-independent generator for german.
\newblock {\em Advances in Natural Language Generation}, 2.

\bibitem[\protect\citename{S{\o}gaard \bgroup et al.\egroup }2015]{sogaard:15}
Anders S{\o}gaard, {\v{Z}}eljko Agi{\'c}, H{\'e}ctor~Mart{\'\i}nez Alonso,
  Barbara Plank, Bernd Bohnet, and Anders Johannsen.
\newblock 2015.
\newblock Inverted indexing for cross-lingual {NLP}.
\newblock In {\em Proc. of ACL-IJCNLP 2015}.

\bibitem[\protect\citename{Srivastava \bgroup et al.\egroup
  }2014]{srivastava:14}
Nitish Srivastava, Geoffrey Hinton, Alex Krizhevsky, Ilya Sutskever, and Ruslan
  Salakhutdinov.
\newblock 2014.
\newblock Dropout: A simple way to prevent neural networks from overfitting.
\newblock {\em Journal of Machine Learning Research}, 15(1):1929--1958.

\bibitem[\protect\citename{Sutskever \bgroup et al.\egroup }2014]{sutskever:14}
Ilya Sutskever, Oriol Vinyals, and Quoc~V. Le.
\newblock 2014.
\newblock Sequence to sequence learning with neural networks.
\newblock In {\em NIPS}.

\bibitem[\protect\citename{T{\"a}ckstr{\"o}m \bgroup et al.\egroup
  }2012]{tackstrom:12}
Oscar T{\"a}ckstr{\"o}m, Ryan McDonald, and Jakob Uszkoreit.
\newblock 2012.
\newblock Cross-lingual word clusters for direct transfer of linguistic
  structure.
\newblock In {\em Proc. of NAACL-HLT}.

\bibitem[\protect\citename{T{\"a}ckstr{\"o}m \bgroup et al.\egroup
  }2013]{tackstrom:13}
Oscar T{\"a}ckstr{\"o}m, Dipanjan Das, Slav Petrov, Ryan McDonald, and Joakim
  Nivre.
\newblock 2013.
\newblock Token and type constraints for cross-lingual part-of-speech tagging.
\newblock {\em Transactions of the Association for Computational Linguistics},
  1:1--12.

\bibitem[\protect\citename{Tiedemann \bgroup et al.\egroup
  }2014]{tiedemann:14b}
J\"org Tiedemann, Zeljko Agic, and Joakim Nivre.
\newblock 2014.
\newblock Treebank translation for cross-lingual parser induction.
\newblock In {\em Proc. of CoNLL}.

\bibitem[\protect\citename{Tiedemann}2015]{tiedemann:15}
J\"org Tiedemann.
\newblock 2015.
\newblock Cross-lingual dependency parsing with universal dependencies and
  predicted {POS} labels.
\newblock In {\em Proc. of DepLing}.

\bibitem[\protect\citename{Titov and Henderson}2007]{titov:07}
Ivan Titov and James Henderson.
\newblock 2007.
\newblock Constituent parsing with incremental sigmoid belief networks.
\newblock In {\em Proc. of ACL}.

\bibitem[\protect\citename{Vilares \bgroup et al.\egroup }2016]{vilares:16}
David Vilares, Carlos G{\'o}mez-Rodr{\'\i}guez, and Miguel~A. Alonso.
\newblock 2016.
\newblock One model, two languages: training bilingual parsers with harmonized
  treebanks.
\newblock arXiv:1507.08449v2.

\bibitem[\protect\citename{Wikipedia}2016]{native:07}
Wikipedia.
\newblock 2016.
\newblock List of languages by number of native speakers.
\newblock \url{http://bit.ly/1LUP5kJ}.
\newblock Accessed: 2016-01-26.

\bibitem[\protect\citename{Xiao and Guo}2014]{xiao:14}
Min Xiao and Yuhong Guo.
\newblock 2014.
\newblock Distributed word representation learning for cross-lingual dependency
  parsing.
\newblock In {\em Proc. of CoNLL}.

\bibitem[\protect\citename{Yarowsky \bgroup et al.\egroup }2001]{yarowsky:01}
David Yarowsky, Grace Ngai, and Richard Wicentowski.
\newblock 2001.
\newblock Inducing multilingual text analysis tools via robust projection
  across aligned corpora.
\newblock In {\em Proc. of HLT}.

\bibitem[\protect\citename{Zeman and Resnik}2008]{zeman:08}
Daniel Zeman and Philip Resnik.
\newblock 2008.
\newblock Cross-language parser adaptation between related languages.
\newblock In {\em Proc. of IJCNLP}.

\bibitem[\protect\citename{Zhang and Barzilay}2015]{zhang:15}
Yuan Zhang and Regina Barzilay.
\newblock 2015.
\newblock Hierarchical low-rank tensors for multilingual transfer parsing.
\newblock In {\em Proc. of EMNLP}.

\end{thebibliography}

\end{document}